\documentclass{article}

\usepackage{arxiv}

\usepackage[utf8]{inputenc} % allow utf-8 input
\usepackage[T1]{fontenc}    % use 8-bit T1 fonts
\usepackage{hyperref}       % hyperlinks
\usepackage{url}            % simple URL typesetting
\usepackage{booktabs}       % professional-quality tables
\usepackage{amsfonts}       % blackboard math symbols
\usepackage{nicefrac}       % compact symbols for 1/2, etc.
\usepackage{microtype}      % microtypography
\usepackage{lipsum}		% Can be removed after putting your text content
\usepackage{graphicx}
\usepackage{natbib}
\usepackage{doi}
\usepackage{amsmath}
\usepackage{caption}
\usepackage{subcaption}

\title{An Integrated Representation \& Compression Scheme Based on Convolutional Autoencoders with 4D-DCT Perceptual Encoding for High Dynamic Range Light Fields}

\author{ \href{https://orcid.org/0000-0001-5262-6724}{\includegraphics[scale=0.06]{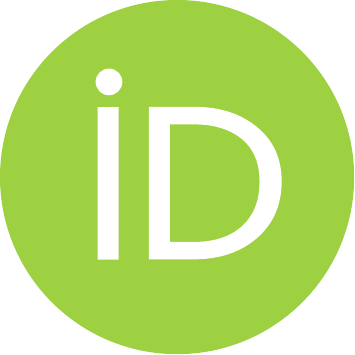}\hspace{1mm}Sally Khaidem}\\
	Department of Electrical Engineering\\
	Indian Institute of Technology Madras\\
	Chennai 600036, India \\
	\texttt{ee20d041@smail.iitm.ac.in} \\
	\And
	\href{https://orcid.org/0000-0003-3243-3321}{\includegraphics[scale=0.06]{orcid.pdf}\hspace{1mm}Mansi Sharma} \\
	Department of Electrical Engineering\\
	Indian Institute of Technology Madras\\
	Chennai 600036, India \\
	\texttt{mansisharma@ee.iitm.ac.in} \\
}

\date{}

\renewcommand{\shorttitle}{4DDCT-UCS}

\hypersetup{
pdftitle={A template for the arxiv style},
pdfsubject={q-bio.NC, q-bio.QM},
pdfauthor={David S.~Hippocampus, Elias D.~Striatum},
pdfkeywords={First keyword, Second keyword, More},
}

\begin{document}
\maketitle

\begin{abstract}
The emerging and existing light field displays are highly capable of realistic presentation of 3D scenes on auto-stereoscopic glasses-free platforms. The light field size is a major drawback while utilising 3D displays and streaming purposes. When a light field is of high dynamic range, the size increases drastically. In this paper, we propose a novel compression algorithm for a high dynamic range light field which yields a perceptually lossless compression. The algorithm exploits the inter and intra view correlations of the HDR light field by interpreting it to be a four-dimension volume. The HDR light field compression is based on a novel 4DDCT-UCS (4D-DCT Uniform Colour Space) algorithm. Additional encoding of 4DDCT-UCS acquired images by HEVC eliminates intra-frame, inter-frame and intrinsic redundancies in HDR light field data. Comparison with state-of-the-art coders like JPEG-XL and HDR video coding algorithm exhibits superior compression performance of the proposed scheme for real-world light fields.
\end{abstract}

% keywords can be removed

\section{Introduction}
\label{sec:introduction}
\begin{figure}[!ht]
\centerline{\includegraphics[width=0.30\textwidth]{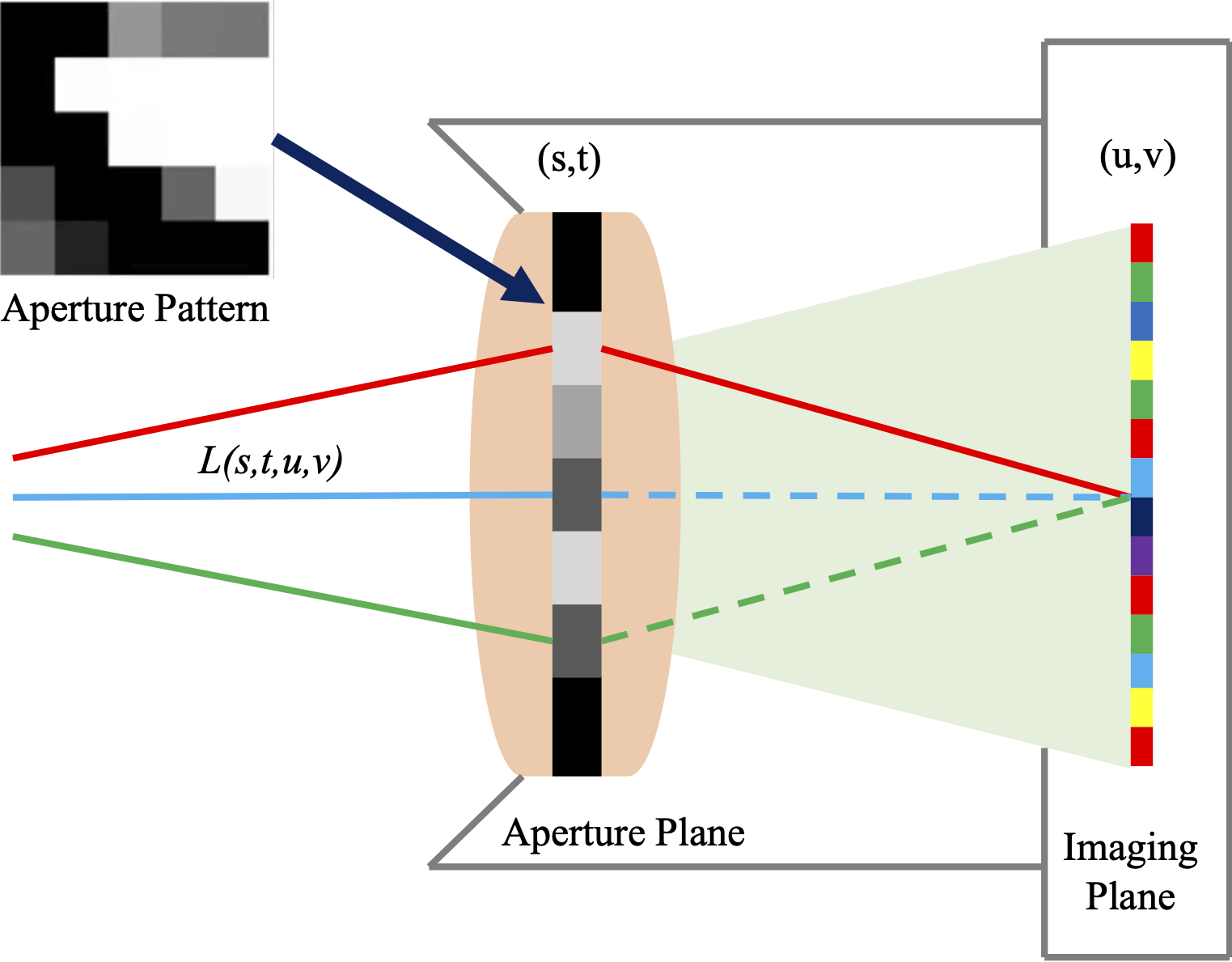}}
\caption{Schematic diagram of a coded aperture camera.}
\label{schematic_ca}
\end{figure}
The light field imaging in the latest generation of auto-stereoscopic and glasses-free displays have revamped their performance in rendering 3D scenes and contents. The support for continuous motion parallax, greater depth-of-field and a wider field-of-view enable computational multi-view light field displays to offer a more realistic viewing experience \cite{1a1}. Imaging of high dynamic range (HDR) light field is relatively an unexplored domain due to the computational expensiveness as compared to a low dynamic range (LDR) light field compression algorithms and lack of hardware that can fully utilise and display the HDR content. There is a growing necessity for efficient representation and coding pipelines for HDR light field data that are suitable for 3D displays and high fidelity streaming applications.

\begin{figure}[!t]
\centerline{\includegraphics[width=1\textwidth]{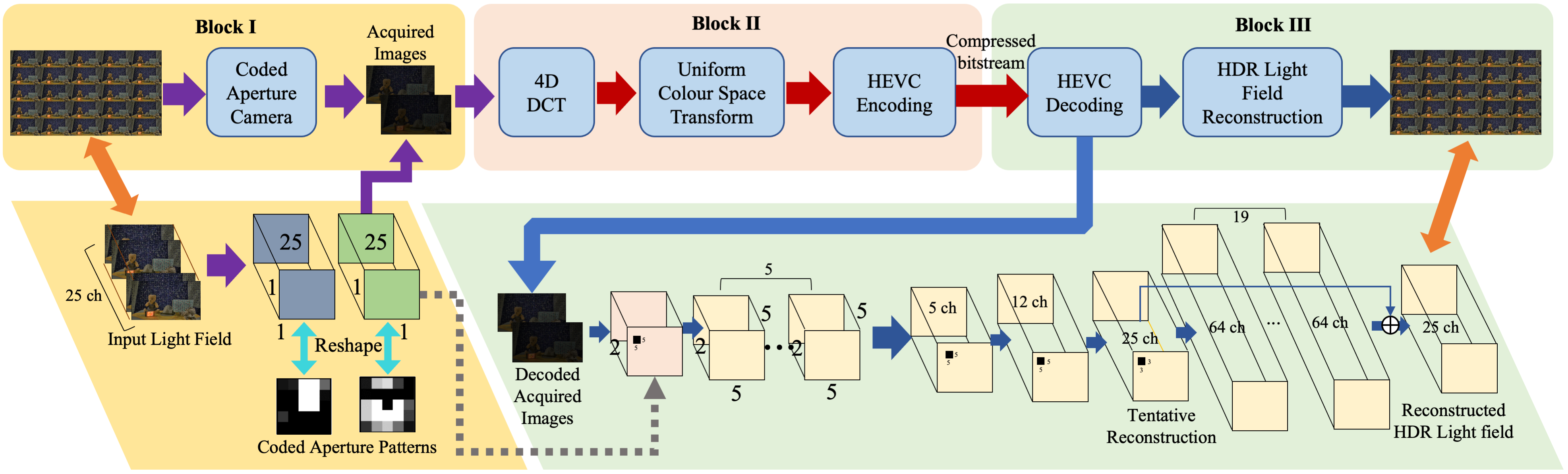}}
\caption{The complete workflow of proposed 4DDCT-UCS scheme.}
\label{workflow}
\end{figure}

This paper focuses on an efficient representation and compression of HDR light field using our novel 4D-DCT Uniform Colour Space (4DDCT-UCS) scheme. The complete workflow of the proposed scheme is illustrated in Fig.~\ref{workflow}. The coded aperture camera shown in Fig.~\ref{schematic_ca} is simulated to obtain acquired images from the HDR light field \cite{1a13,1a14}. The module of acquiring HDR light field is based on an auto-encoder framework \cite{1a37}. The encoder and decoder network are connected and are trained to achieve the best identity mapping between the input and the output. The light field gets reduced to two acquired images through the physical imaging process of a coded aperture camera system. Reconstruction of the light field is performed utilising the acquired images. The 4DDCT-UCS algorithm compresses the acquired images by applying 4D Discrete Cosine Transform (DCT) \cite{1a15} for efficient entropy encoding \cite{1a25}. Further encoding based on perception is performed by linearly scaling of Luma and non-linear scaling Chroma component in the perceptually uniform Intensity, Protan and Tritan (IPT) colour opponent space \cite{1a16}. The proposed 4DDCT-UCS algorithm allows incremental streaming of the acquired image to the receiver at every time instant. High Efficiency Video Coding (HEVC) \cite{1a17} is used to achieve additional encoding by reducing the inter-frame, intra-frame and other intrinsic redundancies present in the light field. The performance of the proposed scheme is being compared with JPEG-XL \cite{1a18} and ours is significantly superior. It is inferred from the experimental results provided in section~\ref{results} that our proposed algorithm is capable of capturing and high-quality reconstruction of HDR light field from a reduced number of acquired images. The main advantages of the proposed algorithm are:

\begin{itemize}
    \item{4DDCT-UCS effectively reduces the inter and intra frame redundancies present in the acquired images by applying 4D DCT and effectively compressing high frequencies and low frequencies separately.}
    
    \item{The proposed scheme works with less number of images instead of working with the entire HDR light field, thus drastically reducing the computational cost.}
    
    \item{4DDCT-UCS can achieve a high PSNR value at lower bit rates and lower total bytes.}
    
\end{itemize}

\section{Proposed work}
In this section, a novel representation and coding of an HDR light field is proposed. The workflow can be divided into three main components. BLOCK I represents the CNN that reduces the light field into acquired images through a physical imaging process of the coded aperture camera. Block II represents the compression of redundant information present in the acquired images through 4D-DCT and perceptual coding the Luma component and linearly scaling the Chroma component for effective compression. In Block III, the decoding of the encoded light field is performed. 
% HEVC is used to perform encoding and decoding of the compressed bit-stream. 
The complete flow of the proposed pipeline is explained as follows.

\subsection{Light Field to Acquired Images}

The 4D light field ${L(u,v,s,t)}$ is defined by the intersection of light into the two parallel planes ${(u,v)}$ and ${(s,t)}$ \cite{1a33}. A coded aperture camera uses the images acquired through various aperture patterns to computationally construct the light field \cite{1a14}. Fig.~\ref{schematic_ca} illustrates the schematic diagram of a coded aperture camera capturing the incoming light rays by adopting a coordinate system, where ${(u,v)}$ and ${(s,t)}$ are imaging plane and aperture plane respectively. The aperture of such camera has a design pattern or mask which controls the admittance of each position and each acquisition. Inagaki et al. \cite{1a14} demonstrate the ability of an optimised CNN to reconstruct the light field entirely from the acquired images. For $N=2$ acquired images, let $r_{n}(s,t)$ be the transmittance at position $(s,t)$ for the $n^{th}$ acquisition ($n=1$ to $N)$. The acquired images are formed as
\begin{equation}
\begin{aligned}
    a_{n}(u,v) & =\iint r_{n}(s,t)L(s,t,u,v)dsdt \\
    & = \iint r_{n}(s,t)l_{s,t}(u,v)dsdt
    \label{eq:1}
\end{aligned}
\end{equation}
Discretising the aperture plane $(s,t)$, the acquired image formed is simplified as
\begin{equation}
    a_{n}(u,v)= \sum_{s,t}r_{n}(s,t)l_{s,t}(u,v)
    \label{eq:2}
\end{equation}
The entire light field reconstruction is equivalent to estimating $M=25$ sub-aperture views from $N=2$ acquired. images $a_{n}(u,v)$. The auto-encoder framework implemented using Chainer \cite{1a38} is trained on samples from 51 light field datasets. 

The mapping function of the coded aperture camera acquisition process~\eqref{eq:2} can be formulated as
\begin{equation}
    \begin{aligned}
    f: X \to Y
    \label{eq:2.1}
    \end{aligned}
\end{equation}
where, $X$ is the vector containing all pixels of $l_{m}(u,v)$ for all $m\in \{1,..,M\}$. The $X$ vector length for an image having dimensions $p\times q$ is $p\times q\times M$. The vector $Y$ represents all the pixel in $a_{n}(u,v)$ for all $n\in \{1,..,N\}$ and the reconstruction mapping is denoted as
\begin{equation}
    \begin{aligned}
    g: Y \to \hat{X}
    \label{eq:2.2}
    \end{aligned}
\end{equation}
The estimate of $X$ is represented by $\hat{X}$. Under the condition that $N\ll M$, the composite mapping function $h=g \circ f$ must be close to identity as possible, where $f$ and $g$ represent the encoder and decoder of the auto-encoder respectively. The objective of the optimization is expressed with square loss error as 
\begin{equation}
    \begin{aligned}
    \underset{f,g}{argmin}|X-\hat{X}|^2=\underset{f,g}{argmin}\sum_{m}\sum_{u,v}|l_m(u,v)-\hat{l}_m(u,v)|^2
    \label{eq:2.3}
    \end{aligned}
\end{equation}
where, $\hat{l}_m(u,v)$ is the estimate of $l_m(u,v)$. The composite mapping $h=g \circ f$ is implemented using 2D convolutional layers. Aperture patterns $r_{n}(u,v)$ are optimised following the parameters learned in $f$ while the reconstruction algorithm is optimised by parameters learned in $g$ over the learning dataset. The acquisition module produces two acquired images from $5\times5$ sub-aperture images of the HDR light field utilising the two optimised aperture patterns.

\subsection{4D forward Discrete Cosine Transform on the Acquired Images}
Block sizes of $8\times8\times8\times8$ are partitioned in the image component. The 2D acquired images are interpreted as a 4D volume. 1D DCT in~\eqref{eq:3} is applied to all the four dimensions of the volume \cite{1a15} as DCT has good decorrelation properties and is less computationally complex \cite{1a27}. The 4D block are transformed from spatial to frequency domain such that the energy is concentrated around the DC coefficients.

\begin{equation}
    \begin{aligned}
        X_k&=\sqrt{\frac{2}{N}}c_k\sum_{n=0}^{N-1}x_n cos\left [ \frac{(2n+1)k\pi}{2N} \right ]\label{eq:3}\\
    c_k&=\left\{\begin{matrix}
    1/\sqrt2 & k=0 \\
    1 & otherwise \\
    \end{matrix}\right. 
    \end{aligned}
\end{equation}

A quantization matrix is applied on the coefficients as denoted in~\eqref{eq:3.1}, where ${X(k)}$ is the DCT coefficient block, ${Q}$ is the quantization matrix, ${X_q(k)}$ is the quantized DCT coefficient block and \textbf{k} is the coefficient index. Uniform quantization is enforced by dividing the coefficients in a block with a constant value. In each 2D slice of the 4D block, the quantization scheme is applied. The quantized DC and AC coefficients are treated separately. 

\begin{equation}
\begin{aligned}
X_q(\textbf{k})=round\left [ \frac{X(\textbf{k})}{Q(\textbf{k})} \right ]\label{eq:3.1}
\end{aligned}
\end{equation}
Run-length encoding is performed to compress the AC coefficients, while Differential Pulse Code Modulation (DPCM) \cite{1a22} is used for the DC coefficient. Finally, the Huffman encoder \cite{1a39} is used to entropy encode both coefficients generating a bitstream. 
   
\begin{equation}
\begin{aligned}
X(\textbf{k})=X_q(\textbf{k})\times Q(\textbf{k})
\label{eq:3.2}
\end{aligned}
\end{equation}

\subsection{Uniform Colour Space Transform on DCT compressed Acquired Images}
 The 4D DCT compressed acquired images are transformed into perceptually uniform Intensity, Protan and Tritan (IPT) colour opponent space \cite{1a16}. The intensity information is extracted, linearly scaled \cite{1a19} and encoded by a Perceptual Transfer Function (PTF). An error minimising function (EMF) encode the non-linear scaled chroma information to minimise the frequently encountered quantization error. The PTF is designed following recommendations of REC 1886 \cite{1a20} taking into account three intensity ranges (low, mid and high intensity) such that $f(\cdot)$ : $I'\to L$ 

\begin{equation}
    \begin{aligned}
    L=\left\{\begin{matrix}
    a.I' & if & I'<I'_s; \\
    b.I'{^{\frac{1}{c}}}+d & if & I'\in [I'_s,I'_p); \\
    e.\log_{10}(I')+f & if & I'\in [I'_p,I'_h]; \\
    \end{matrix}\right.
    \end{aligned}
    \label{eq:6}
\end{equation}

Similarly, the inverse function, $f^{-1}(\cdot)$ is formulated as 

\begin{equation}
    \begin{aligned}
    I'=\left\{\begin{matrix}
    \frac{L}{a} & if & L < L_s; \\
    \left (\frac{L-d}{b}\right )c & if & L\in [L_s,L_p); \\
    10^{\left(\frac{L-f}{e}\right)} & if & L\in [L_p,L_h]; \\
    \end{matrix}\right.
    \end{aligned}
    \label{eq:7}
\end{equation}
The power value ($\lambda$) is appropriately derived so that quantization error due to discretisation is minimised when applied to the input chroma information. For a target 10-bit depth, the discretised output channel $P_{out}\in(0,1023]$ can be optimally obtained from input channel $P_{in}\in(0,1]$ using~\eqref{eq:8}.
\begin{equation}
    \begin{aligned}
    P_{out} = \left \lfloor (P_{in})^\lambda \cdot (2^n-1) \right \rfloor
    \end{aligned}
    \label{eq:8}
\end{equation}
The power function $\lambda$ is computed as shown in~\eqref{eq:9}, where $M,N$ are the horizontal and vertical resolution of the frame respectively.
\begin{equation}
    \begin{aligned}
    \resizebox{0.40\textwidth}{!}{%
    $\underset{\lambda}{argmin}\left ( \frac{1}{MN}\sum_{j=1}^{N}\sum_{i=1}^{M}\left| \left ( \frac{ \left \lfloor (P_in)^\lambda \cdot (2^n-1) \right \rfloor}{(2^n-1)} \right)^{\frac{1}{\lambda}} - P_{in} \right| \right )$%
    }
    \end{aligned}
    \label{eq:9}
\end{equation}
Metadata containing crucial information like scaling data, maximum and minimum chroma values, etc. are generated  for each frame by the proposed scheme and stored as a lookup table (LUT). The final bitstream is encoded using HEVC.

\subsection{Reconstruction of HDR Light Field from Decoded 4DDCT-UCS Acquired Images}
The 4DDCT-UCS acquired images are decoded from the primary bitstream and the secondary metadata stream. The auto-encoder network shown in Fig.~\ref{workflow} is used to reconstruct the HDR light field from the decompressed acquired images. Hence, we achieve a complete scheme capable of efficient representation and compression of HDR light field.

\section{Results and Analysis}
\label{results}
\begin{figure}[!ht]
\centering
    \begin{subfigure}{0.40\textwidth}
      \includegraphics[width=\linewidth]{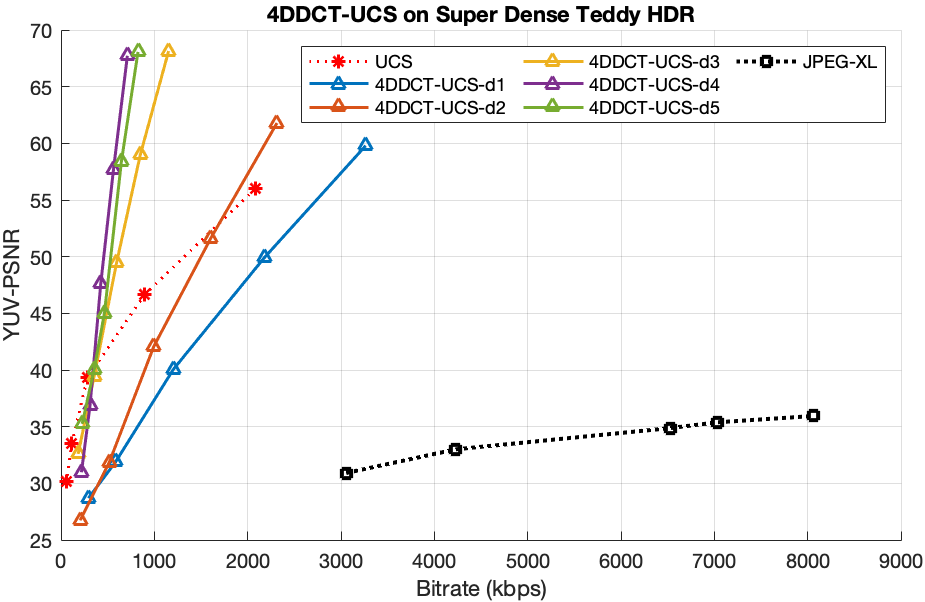}
      \caption{}
      \label{yuv_psnr}
    \end{subfigure} 
    \begin{subfigure}{0.40\textwidth}
      \includegraphics[width=\linewidth]{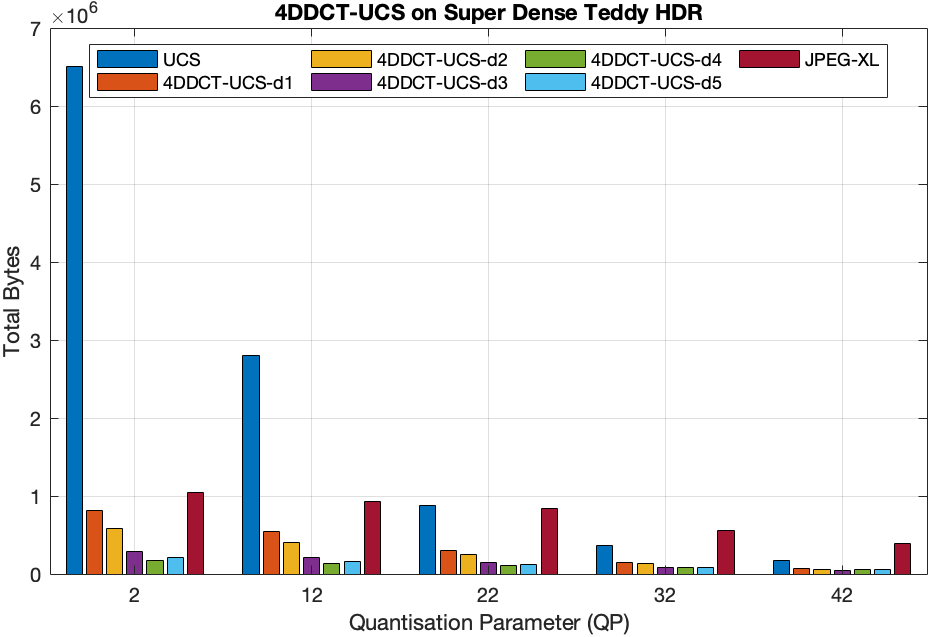}
      \caption{}
      \label{bytes_written}
    \end{subfigure} 
\caption{\footnotesize~\textbf{(a)}: Rate-distortion curves for the proposed 4DDCT-UCS scheme, UCS and JPEG-XL for the HDR light field \textit{Super Dense Teddy}. \textbf{(b)}: Total bytes for the proposed 4DDCT-UCS scheme, UCS and JPEG-XL for the HDR light field \textit{Super Dense Teddy}}
% \label{yuv_psnr}
\end{figure}
% Currently, there is only a handful of real HDR light field datasets. 
Evaluation of the proposed pipeline 4DDCT-UCS is conducted on a super-dense HDR light field `\textit{Super Dense Teddy}' \cite{1a21} dataset having an angular resolution of $50\times50$ views and spatial resolution of $5168\times3448$ pixels. The proposed scheme was implemented on a single high-end HP OMEN X 15-DG0018TX system with 9th Gen i7-9750H, 16 GB RAM, RTX 2080 8 GB Graphics, and Windows 10 operating system. Sample sets of $64\times64$ pixels 2D image blocks are extracted from the same location of the light field datasets for training the network for 20 epochs with a batch size of 15.  

\begin{figure}[t]
\centering
    \begin{subfigure}{0.23\textwidth}
      \includegraphics[width=\linewidth]{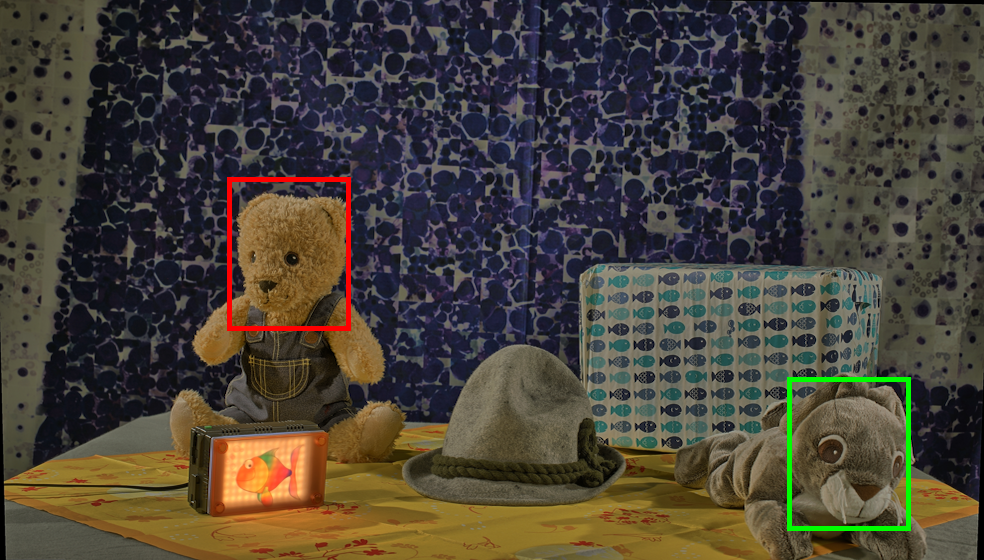}
    \end{subfigure} 
    \begin{subfigure}{0.23\textwidth}
      \includegraphics[width=\linewidth]{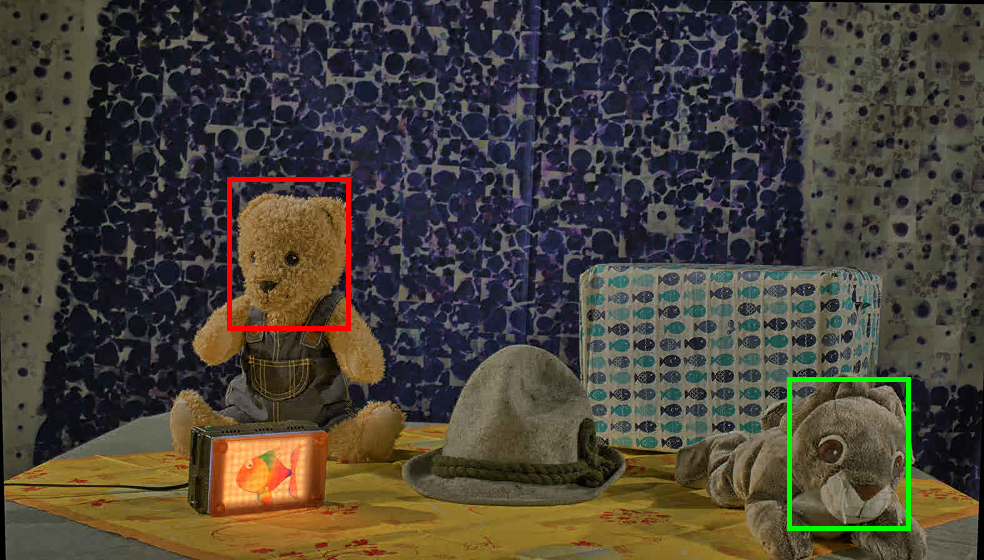}
    \end{subfigure} 
    \begin{subfigure}{0.23\textwidth}
      \includegraphics[width=\linewidth]{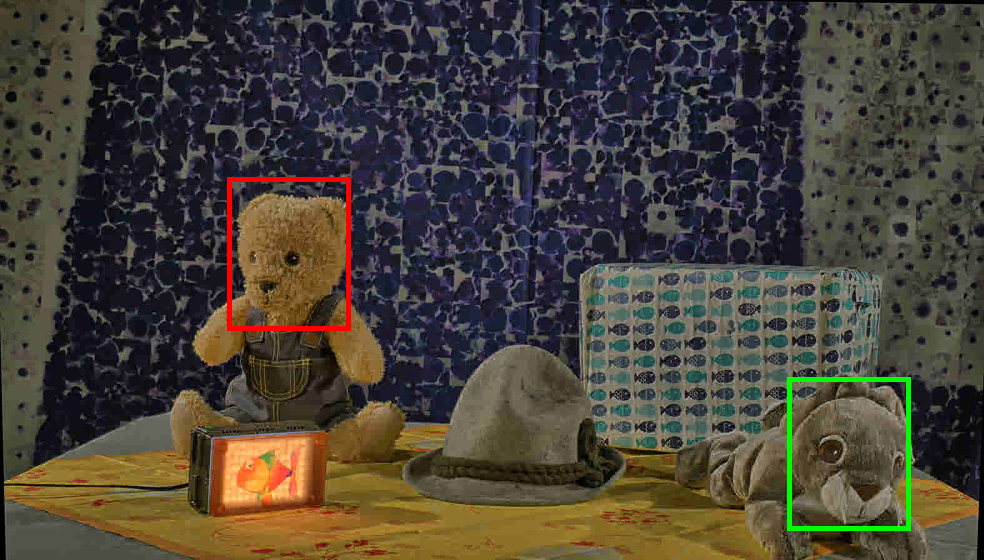}
    \end{subfigure}
    \begin{subfigure}{0.23\textwidth}
      \includegraphics[width=\linewidth]{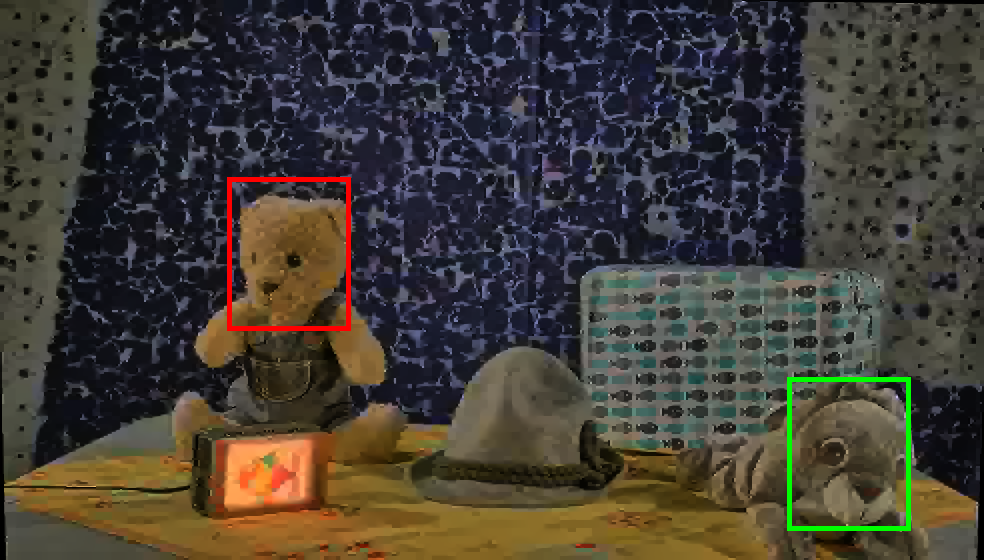}
    \end{subfigure}\\
    \begin{subfigure}{0.23\textwidth}
    \centering
        \begin{subfigure}{0.48\textwidth}
          \includegraphics[width=\linewidth]{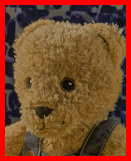}  
        \end{subfigure}
        \begin{subfigure}{0.48\textwidth}
          \includegraphics[width=\linewidth]{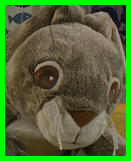}  
        \end{subfigure}
    \end{subfigure} 
    \begin{subfigure}{0.23\textwidth}
    \centering
        \begin{subfigure}{0.48\textwidth}
          \includegraphics[width=\linewidth]{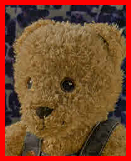}  
        \end{subfigure}
        \begin{subfigure}{0.48\textwidth}
          \includegraphics[width=\linewidth]{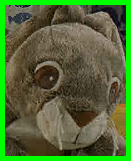}  
        \end{subfigure}
    \end{subfigure} 
    \begin{subfigure}{0.23\textwidth}
    \centering
        \begin{subfigure}{0.48\textwidth}
          \includegraphics[width=\linewidth]{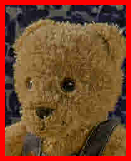}  
        \end{subfigure}
        \begin{subfigure}{0.48\textwidth}
          \includegraphics[width=\linewidth]{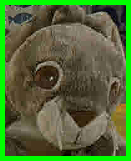}  
        \end{subfigure}
    \end{subfigure} 
    \begin{subfigure}{0.23\textwidth}
    \centering
        \begin{subfigure}{0.48\textwidth}
          \includegraphics[width=\linewidth]{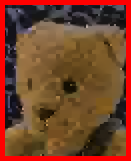}  
        \end{subfigure}
        \begin{subfigure}{0.48\textwidth}
          \includegraphics[width=\linewidth]{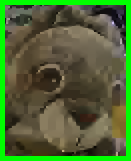}  
        \end{subfigure}
    \end{subfigure} \\
    \begin{subfigure}{0.23\textwidth}
      \includegraphics[width=\linewidth]{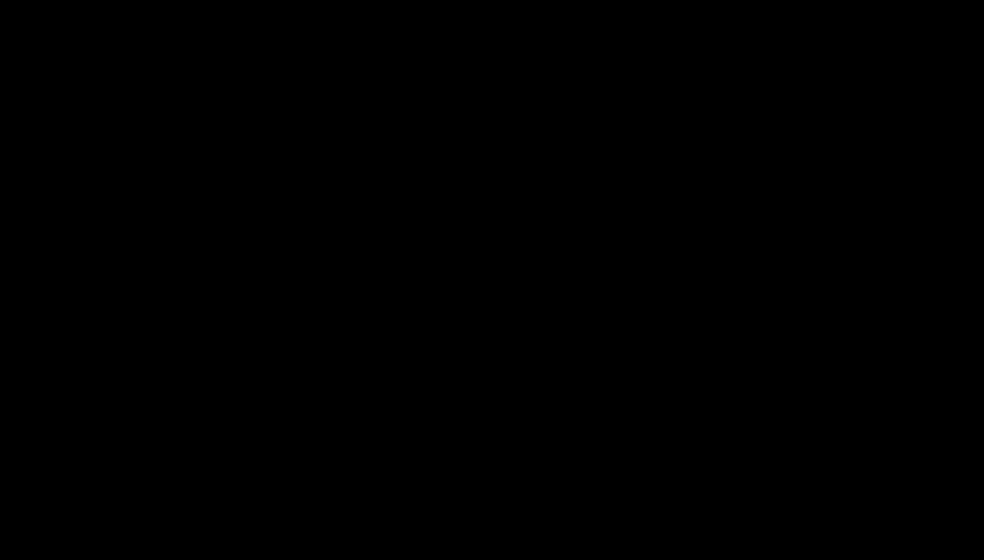}
      \caption{\footnotesize~Ground Truth }
    \end{subfigure} 
    \begin{subfigure}{0.23\textwidth}
      \includegraphics[width=\linewidth]{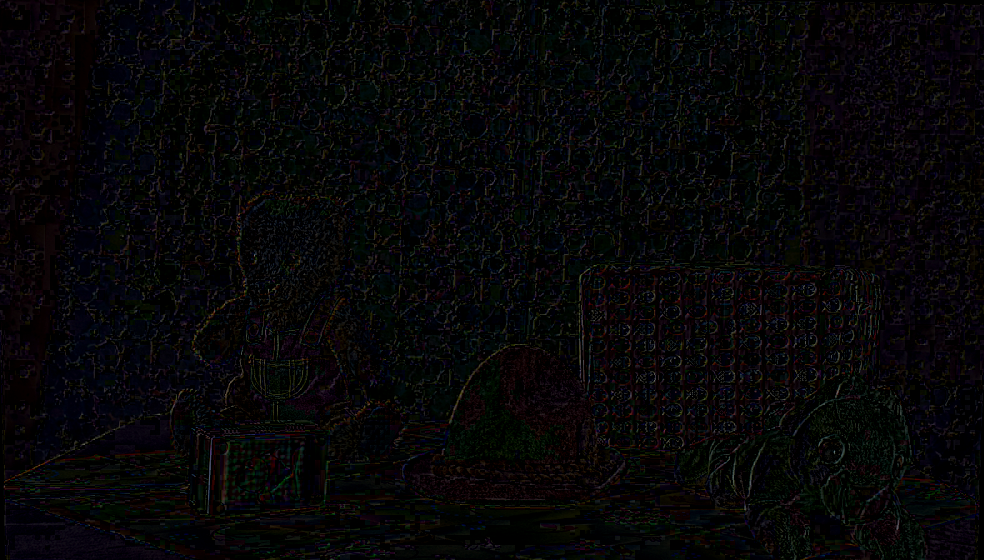}
      \caption{\footnotesize~Ours }
    \end{subfigure} 
    \begin{subfigure}{0.23\textwidth}
      \includegraphics[width=\linewidth]{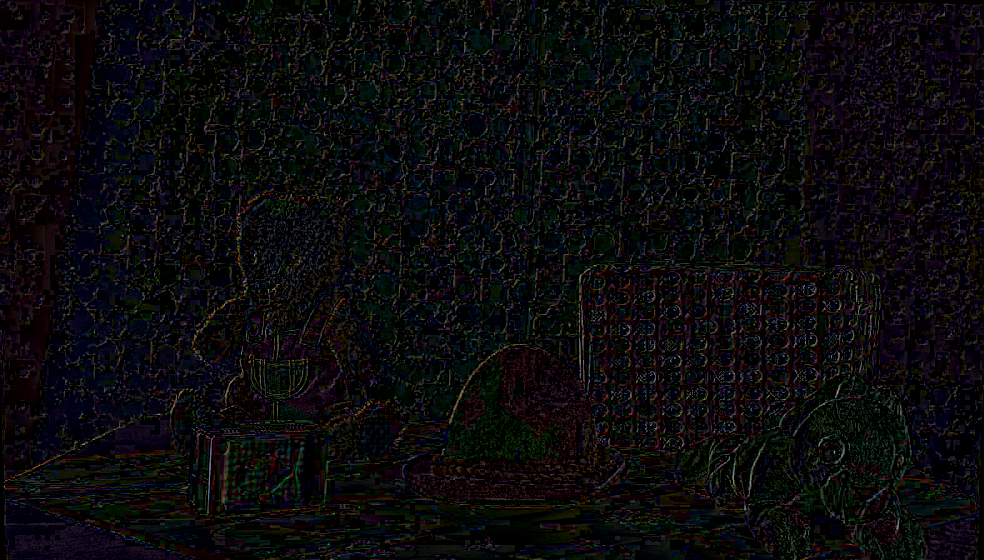}
      \caption{\footnotesize~UCS }
    \end{subfigure}
    \begin{subfigure}{0.23\textwidth}
      \includegraphics[width=\linewidth]{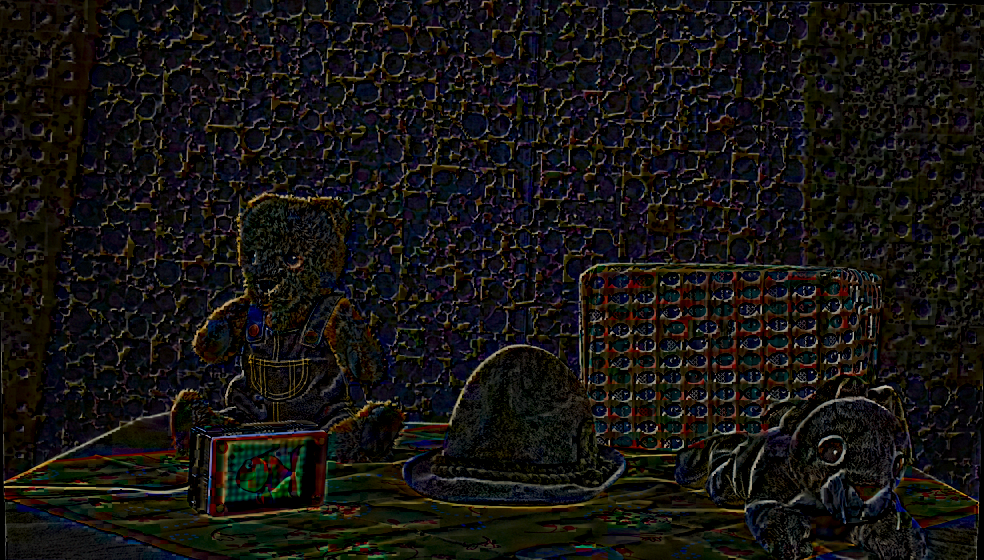}
      \caption{\footnotesize~JPEG-XL }
    \end{subfigure}
\caption{\footnotesize~Central views of the reconstructed light field and difference from ground truth. $First\ row$: Central views. $Second\ row$: Magnified regions of the central views. $Third\ row$: Difference of reconstructed central view from the ground truth.}
\label{GT_difference}
\end{figure}

The distortion parameter `$d$' gives us the option of high compressibility or high quality of the data. The acquired images are compressed with varying distortion levels starting from $d=1$ to $d=5$. The overall bitstream is encoded by HEVC. The profile used is Main10 \cite{1a31} which is designed for encoding streams of 10-bit depth and 4:2:0 chroma subsampling \cite{1a32}. The quantization parameters (QP) is set as $2,12,22,32,42$ covering most of the HEVC compression range. The decoding and reconstruction of the HDR light field are also performed.

\subsection{Comparative Analysis}
The performance of the proposed scheme 4DDCT-UCS is compared with JPEG-XL \cite{1a18} and UCS \cite{1a19}. In Fig~\ref{yuv_psnr}, the analysis of the comparison in terms of rate-distortion is depicted. We observe from the plot that our proposed scheme exhibits high fidelity compression with a lower bit rate as compared to JPEG-XL and UCS. Since the number of input images in our proposed processing pipeline is only two, the total bytes is very low compared to the other two algorithms as shown in Fig.~\ref{bytes_written}. The reconstructed images, along with their difference from the ground truth, are presented in the Fig.~\ref{GT_difference}. Further, an objective assessment using Bjontegaard~\cite{bjontegaard2001calculation} metric to compare bitrate reduction of proposed scheme with respect to JPEG-XL codec and UCS is performed. For the chosen distortion level $d$, the average percent difference in rate change is estimated over a range of quantization parameters. Table~\ref{bdrate} depicts comparative BD-PSNR and BD-Rate. Thus, the proposed scheme can reconstruct the light field from just two acquired images with a minor deviation from the original light field.
% Table generated by Excel2LaTeX from sheet 'Sheet2'
\begin{table}[!ht]
    \caption{\footnotesize Bjontegaard percentage rate savings for the proposed compression Scheme (4DDCT-UCS) with respect to JPEG-XL and UCS on Super Dense Teddy HDR light field.}
    \centering
    \begin{tabular}{ccccc}
    \multicolumn{1}{c}{} & \multicolumn{2}{c}{\textbf{JPEG-XL}} & \multicolumn{2}{c}{\textbf{UCS}} \\
        \hline
          \textbf{Distortion(\textit{d})}& \textbf{BD-Rate} & \textbf{BD-PSNR} & \textbf{BD-Rate} & \textbf{BD-PSNR} \\
          \hline
          1 & -86.201 & 27.668 & 193.58 & -10.326\\
          2 & -88.119 & 36.119 & 119.51 & -7.5537\\
          3 & -96.058 & 61.3 &	-21.767 & 3.9596\\
          4 & -94.651 & 66.725 & -24.748 & 5.1009\\
          5 & -96.852 & 83.711 & -35.6 & 4.9842\\
          \hline
    \end{tabular}%
  \label{bdrate}%
\end{table}%
\section{Conclusion}
A novel representation and perceptual coding scheme for HDR light field based on convolutional autoencoders and 4D-DCT are presented in this paper. The proposed scheme exhibits adaptable bit rate support which is crucial for adjusting compression levels and quality according to the available bandwidth. It also provides an end-to-end framework, where it can capture and reconstruct the light field back. The proposed algorithm outperforms the state-of-the-art coder JPEG-XL and UCS. In future work, we aim to incorporate a deep-learning based perceptual transfer function into the algorithm. We also aim to verify our simulated results on a light field display. 

\bibliographystyle{unsrtnat}
\bibliography{root} 
\end{document}